\documentclass{article}

\usepackage[final,nonatbib]{neurips_2021_ml4ps}
\usepackage[utf8]{inputenc} 
\usepackage[T1]{fontenc}    
\usepackage{hyperref}       
\usepackage{url}            
\usepackage{booktabs}       
\usepackage{amsmath}
\usepackage{amsfonts}       
\usepackage{nicefrac}       
\usepackage{microtype}      
\usepackage{xcolor}         
\usepackage{graphicx}
\usepackage{caption}
\usepackage{subcaption}
\graphicspath{{./}{./image}{./image/results}}

\definecolor{vertfonce}{RGB}{0,127,0}
\definecolor{jaunefonce}{RGB}{246,246,0}
\definecolor{gris25}{gray}{0.80}

\newcommand{\lucas}[1]{{\color{green}\small #1}}

\newcommand{\hidden}[1]{}

\usepackage{biblatex} 
\title{Deep Surrogate for Direct Time Fluid Dynamics}
\author{%
  Lucas ~Meyer\\
  EDF Lab Paris-Saclay, Univ. Grenoble Alpes, Inria, CNRS, Grenoble INP, LIG\\
  \texttt{lucas.meyer@inria.fr}\\
  \And
  Louen ~Pottier\\
  EDF Lab Paris-Saclay, ENS Paris-Saclay\\
  \texttt{louen.pottier@ens-paris-saclay.fr}\\
  \And
  Alejandro ~Ribes\\
  EDF Lab Paris-Saclay\\
  91120 Palaiseau, France\\
  \texttt{alejandro.ribes@edf.fr}
  \And
  Bruno ~Raffin\\
Univ. Grenoble Alpes, Inria, CNRS, Grenoble INP, LIG\\
  38000 Grenoble, France\\
  \texttt{bruno.raffin@inria.fr}\\
}
\date{September 2021}

\begin{document}

\maketitle
\begin{abstract}

The ubiquity of fluids in the physical world explains the need to accurately
simulate their dynamics for many scientific and engineering applications.
Traditionally, well established but resource intensive CFD solvers provide such simulations. The recent years have seen a surge of deep learning surrogate models substituting these solvers to alleviate the simulation process. Some approaches to build data-driven surrogates mimic the solver iterative process. They infer the next state of the fluid given its previous one. Others directly infer the state from time input. Approaches also differ in their management of the spatial information. Graph Neural Networks (GNN) can address the specificity of the irregular meshes commonly used in CFD simulations. In this article, we present our on-going work to design a novel direct time GNN architecture for irregular meshes. It consists of a succession of graphs of increasing size connected by spline convolutions. We test our architecture on the Von K\'arm\'an’s vortex street benchmark. It achieves small generalization errors while mitigating error accumulation along the trajectory.

\end{abstract}

\section{Introduction}
Computational Fluid
Dynamics (CFD) solvers have benefited from strong
developments for decades, being critical for many scientific and industrial
applications. Eulerian based CFD solvers rely on a discretization of the
simulation space, i.e. a mesh, augmented with different fields like velocity and
pressure, and constrained by initial and boundary conditions. The solver
progresses by discrete time steps, building from the state $u_t$ at time $t$, a
new state $u_{t+\delta t}$ compliant with the Navier-Stokes equations. This
process is compute intensive and often requires supercomputers for industrial
grade simulations.

Connecting CFD with machine learning has received a renewed attention with the
emergence of deep learning \cite{stevens2020ai,brunton2020machine}. The goal is
often to augment or supplant classical solvers for improved performance in terms
of compute speed, error, and resolution. In this paper, we focus on {\it Deep
Surrogates} where a neural network is trained to provide a quality solution to the
Navier-Stokes equations for a given domain, initial and boundary conditions.

Deep surrogates are currently addressed through different approaches. Data-free
surrogates inject into the loss the different terms of the Navier-Stokes
equations to comply with, leveraging automatic differentiation to compute the
necessary derivatives. Data-driven surrogates train from the data produced by a
traditional CFD solver. The surrogate can mimic the solver iterative process,
being trained to compute $u_{t+\delta t}$ from $u_t$. But as the fluid
trajectories are available at training time, other surrogates are trained to
directly produce $u_t$ from the parameters characterizing the conditions at $t$.
Surrogates also differ in their approach to space discretization. If the mesh is
a regular grid, CNNs can be used. Irregular meshes or particle based approaches
are more challenging, and can be addressed through some variations of GNNs
\cite{bronstein2017geometric,battaglia2018relational}. These various
approaches result in different trade-offs and limitations regarding precision,
generalization capabilities, and scalability.

This paper presents our on-going work to design a surrogate for industrial and
scientific CFD applications. The resulting surrogate is expected to be used for
interactive data analysis, sensibility analysis, and digital twins. Precision
and support for large irregular meshes are a priority rather than wide
generalization capabilities as often targeted, for instance, by surrogates for
computer graphics applications. Iterative surrogates tend to lose precision on
long trajectories due to error accumulation. We thus opt for a direct time
approach. But so far direct time surrogates have been mainly studied at small
scale with regular meshes. Our contribution is a novel neural architecture that
morphes the initial state parameters at $t$ into the final irregular mesh
providing $u_t$ through a succession of layers of increasing size based on
spline convolutions \cite{fey2018splinecnn}. Early experiments with the Von
K\'arm\'an’s vortex street benchmark show that our architecture achieves small
generalization errors (RMSE at about $10^{-2}$) not subject to error
accumulation along the trajectory.

\section{Related Work}
The seminal work of Raissi et al. introduces physics informed neural networks
(PINNs) \cite{raissi2019physics}. These data-free surrogate models are trained
by minimizing the residual of the underlying Partial Differential Equation (PDE), whose terms are computed by
automatic differentiation. PINNs have then been declined and applied to PDEs
similar to the Navier-Stokes equations
\cite{jin2021nsfnets,wandel2021learning,donon2020deep,kharazmi2021hp}. In spite
of their elegant data-free approach, PINNs are still unable to match traditional
solver accuracy \cite{cai2021physics}. Whereas PINNs must learn representations
of space and time to solve the PDE, other approaches guide further the model on
how to treat space and time. \hidden{Regarding time, we classify these
approaches as either iterative or direct time prediction. Regarding space, we
distinguish regular grid and mesh based models.}

Iterative methods reconstruct the full dynamics by taking prediction $u_t$ of
the previous time step as input for computing the next one $u_{t+\delta t}$
\cite{kim2019deep,sanchez2020learning,pfaff2020learning,wandel2021learning}.
Inputs and outputs are not necessarily limited to one time step
\cite{wiewel2019latent}. Data-driven iterative methods can only reproduce
dynamics with the same time step $\delta t$ as seen during training. Sanchez et al.
overcome this limitation by mixing predictions of the derivative $u_{t+\delta t}
- u_t$ and traditional numerical methods \cite{sanchez2019hamiltonian}. Liu et
al. address this problem by training separately identical networks on different
step sizes \cite{liu2020hierarchical}. Iterative methods present a more serious
problem: error accumulation. Each iteration of the surrogate model generates
error, which accumulates along the simulation and eventually hinders its
stability. In \cite{sanchez2020learning,pfaff2020learning}, the authors add
noise to the inputs to train the model to compensate its own error accumulation.
Nonetheless, the initial noise level has to be set empirically. Traditional iterative
solvers are also subject to error accumulation but with errors bounded by
theoretical guarantees \cite{glowinski2003finite}.

Instead of relying on intermediate steps, direct time prediction models predict
$u_t$ directly from the input time $t$. Consequently, the error is independent
from other time steps. PINNs and their extensions provide examples of direct
time predictions \cite{raissi2019physics,lu2019deeponet}. The common approach,
as found in \cite{sirignano2018dgm}, is to predict $u(x,t)$ from $x$ and $t$
representing a unique point located in space and time. Most of direct time
prediction models are mesh-free. They do not take advantage of spatial
correlations between inputs. To our knowledge, only \cite{harsch2021direct}
proposes direct time predictions on meshes. However, this surrogate model
predicts steady flows, which by definition are flows that do not depend on time.

The meshes found in traditional solvers can also be used by surrogate models. In
case they are regular, each time step can be seen as an image. This enables the
use of traditional deep learning algorithms for image analysis tasks to predict
$u_t$. For instance U-Net architectures are employed in
\cite{thuerey2020deep,wang2020towards}, or auto-encoders in \cite{kim2019deep}.
At inference, these surrogate models are constrained by the size of the grid
used during training. Conversely, models trained with irregular meshes adapt
more easily to different resolutions. For instance, in \cite{pfaff2020learning}
the network is trained with irregular meshes of coarse resolution and generalizes
well to meshes of finer resolutions. Furthermore, irregular meshes are commonly
used in CFD algorithms \cite{ferziger2002computational}. It is thus natural to
find models trained with simulation data, such as \cite{donon2020deep}, or
mimicking traditional solvers \cite{lino2021simulating}, to be based on
irregular meshes.

The long-time study of fluid dynamics provides physical knowledge that can be
used as inductive bias to train a surrogate model. For instance, PINNs rely
exclusively on the Navier-Stokes equations \cite{raissi2019physics}. One
critical element of these equations in the case of incompressible fluids is the
incompressibility property itself. It translates as a divergence free velocity.
To enforce this property is not trivial, even for traditional methods
\cite{ferziger2002computational}. In \cite{kim2019deep} divergence free velocity
fields are obtained by returning the curl of the network prediction. The
incompressibility is also associated with pressure information traveling at
infinite speed. In a graph based approach this means pressure information must
be constantly shared between all the nodes, which would require as many message
passing as the diameter of the graph. This would become quickly prohibitive for
large meshes. Hence, \cite{lino2021simulating} proposes a hierarchical network
to let information flow across the whole mesh. We retain in our model this
hierarchical approach to bring inductive bias.

\hidden{
As our network relies on spline convolution \cite{Fey-SplineCNN-May2018}, we
should mention the work of Sun et al. in a similar context but with different
goal \cite{sun2021physics}. Whilst they use spline convolutions to discover the
underlying equations of dynamical systems observed through sparse noisy data, we
use such layers to simulate dynamics from known equations.
}

\section{Method}
\begin{figure}
  \centering
  \includegraphics[width=.7\textwidth]{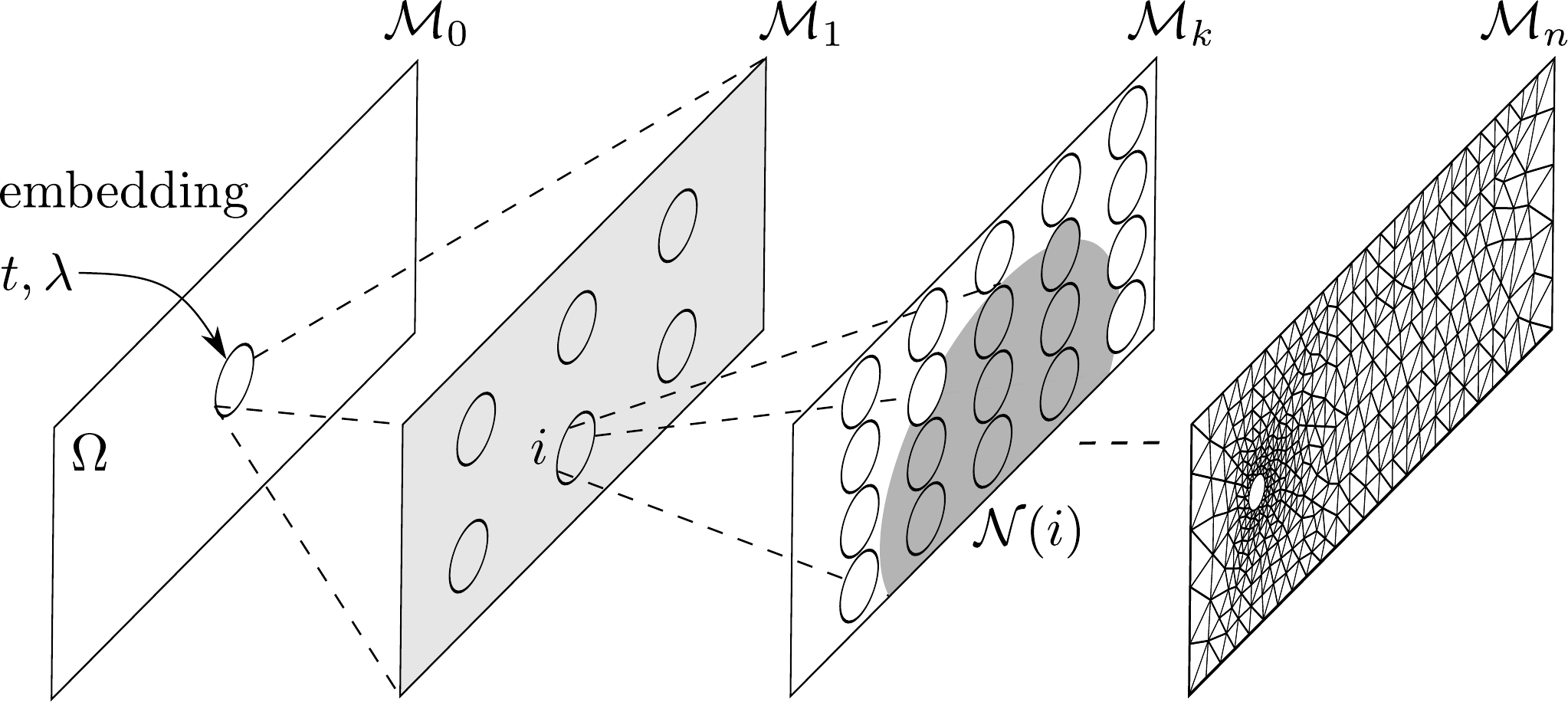}
  \caption{Architecture overview. A  succession of regular meshes
    $\mathcal{M}_k$ defined over the domain $\Omega$, starts with the singleton 
    mesh $\mathcal{M}_0$ embedding the input values $t, \lambda$, and
    ends with the output $\mathcal{M}_n$ corresponding to the simulation mesh. Each node $i$ contributes through
    convolutional operations to the nodes of mesh $\mathcal{M}_k$ that fall in  its neighbourhood $\mathcal{N}(i)$.}
  \label{fig:hierarchical_meshes}
\end{figure}

We propose a {\it data-driven direct time prediction deep surrogate model}
(Fig.~\ref{fig:hierarchical_meshes}). Our deep neural network takes as input a
time step $t$, and one or several physical parameters $\lambda$, which in the
case of CFD represent fluid properties and/or initial conditions. The output of
the network is a prediction $\hat{u}_t$, corresponding to the velocity and
pressure fields of the fluid. This means that the network outputs all the values
of a discretized field. Training is achieved with backprogation of the loss
$\mathcal{L}$, which corresponds to the mean square error relative to the
original simulation $u_t$: $\mathcal{L}=\|\hat{u}_t - u_t \|^2_2$.

The first key point in our approach is that the network is built by use of a mesh $\mathcal{M}(V, X)$, where $V$ is the set of nodes of the mesh and $X$ is the set of their spatial coordinates in the domain $\Omega$. The mesh $\mathcal{M}$ corresponds to the spatial discretization used by the solver to generate the velocity and pressure fields. Thus, our deep surrogate outputs discretized fields (arrays of values) of the pressure and velocity, which are spatially localized by means of this mesh. Our aim is to integrate in the network architecture the geometrical information contained in $\mathcal{M}(V, X)$. For this, we  build our network as a hierarchical graph neural network, i.e. a succession of graphs $\mathcal{M}_k(V^k,X^k, F^k)_{0 \leq k \leq n}$, where $F^k$ denotes the set of feature vectors associated to the spatially localized nodes $V^k$. The last graph of the network $\mathcal{M}_n(V^n, X^n, F^n)$ is fixed by assigning $(V^n, X^n) \equiv (V, X)$ and $F^n$ to the velocity and pressure field predictions. The preceding graphs $\mathcal{M}_{k < n}$ are defined as meshes of gradually thinner resolution starting from $\mathcal{M}_0(V^0, X^0, F^0)$, which is reduced to a singleton associated to an embedding of the inputs $t$ and $\lambda$. The global architecture can be seen as a decoder that predicts the velocity and pressure from latent information about the time
step and physical parameters, by means of progressively refined meshes.

The second key point is how consecutive graphs of our architecture are connected. We define a distance based neighborhood between meshes $\mathcal{M}_k$ and $\mathcal{M}_{k+1}$  as: 
\begin{equation} 
  \forall j \in V^{k+1},\ \mathcal{N}(j) = \{i \in V^{k}, \|x_i - x_j\|_2 \leq r_k\}.
  \label{eq:neighbourhood}
\end{equation}

 This reciprocally defines the neighborhood of any node $i \in V^{k}$ in $V^{k+1}$ as displayed in Figure \ref{fig:hierarchical_meshes}. Once this neighborhood is established, we build a non-regular convolution operator using the spline convolution defined in \cite{fey2018splinecnn}. Thus we obtain a network allowing to deal with non-regular meshes of any dimension, which are the ones usually used in numerical simulations. Figure \ref{fig:hierarchical_meshes} highlights the hierarchical graphs used in the network architecture, which have a direct interpretation in the framework of graph neural networks \cite{battaglia2018relational}. The computation of features of $\mathcal{M}_{k+1}$ is achieved through a learnable node update function that takes as inputs the features of $\mathcal{M}_k$. Indeed, the feature vectors are computed as:

\begin{equation}
  \forall j \in V^{k+1},\ \textbf{f}_j = \frac{1}{|\mathcal{N}(j)|} \sum_{i \in \mathcal{N}(j)}\textbf{H}_\theta(d_{ij})^\top \textbf{f}_i 
  \label{eq:spline_conv}
\end{equation}

where the neighborhood $\mathcal{N}(j)$ is defined using Equation
\ref{eq:neighbourhood}. The kernel $\textbf{H}_\theta(d_{ij})$ is computed by
interpolating the shifting vector $d_{ij}$ between nodes $i$ and $j$ on $m$ control
points with B-splines functions of degree $n$ along the $d$ dimensions of
$\Omega$. Each basis function of the interpolation map is associated to a
trainable weight. We refer the reader to the original article for a more
detailed explanation about the spline convolution \cite{fey2018splinecnn}.

Our network consists in two fully connected layers that embed the parameters $t$
and $\lambda$ in a feature vector $\textbf{f}_0$. Then a total of $n$ spline
convolution layers update nodes of $\mathcal{M}_{k+1}$ from values of
$\mathcal{M}_k$. Each except the last convolution is followed by batch
normalization and ReLU activation. In practice, we set $\mathcal{M}_{k < n}$ as
regular meshes whose nodes are spread over $\Omega$. We set $r_0 = + \infty$ and
$r_k = \sqrt{2}h_k$, where $h_k$ is the minimum distance between nodes of $M_k$.
This allows each node of $\mathcal{M}_n$ to be parented to $\mathcal{M}_0$.  The
intuition is to share information between all the nodes of the output mesh to
support the elliptic character of pressure information. The Navier-Stokes
equations for an unsteady and incompressible fluid are incompletely parabolic
with pressure information travelling at infinite speed across the mesh
\cite{ferziger2002computational}.
\section{Experiment}
\label{sec:experiment}

\begin{figure}[h!]
  \centering
  \begin{subfigure}[c]{0.45\textwidth}
    \centering
    \includegraphics[width=\textwidth]{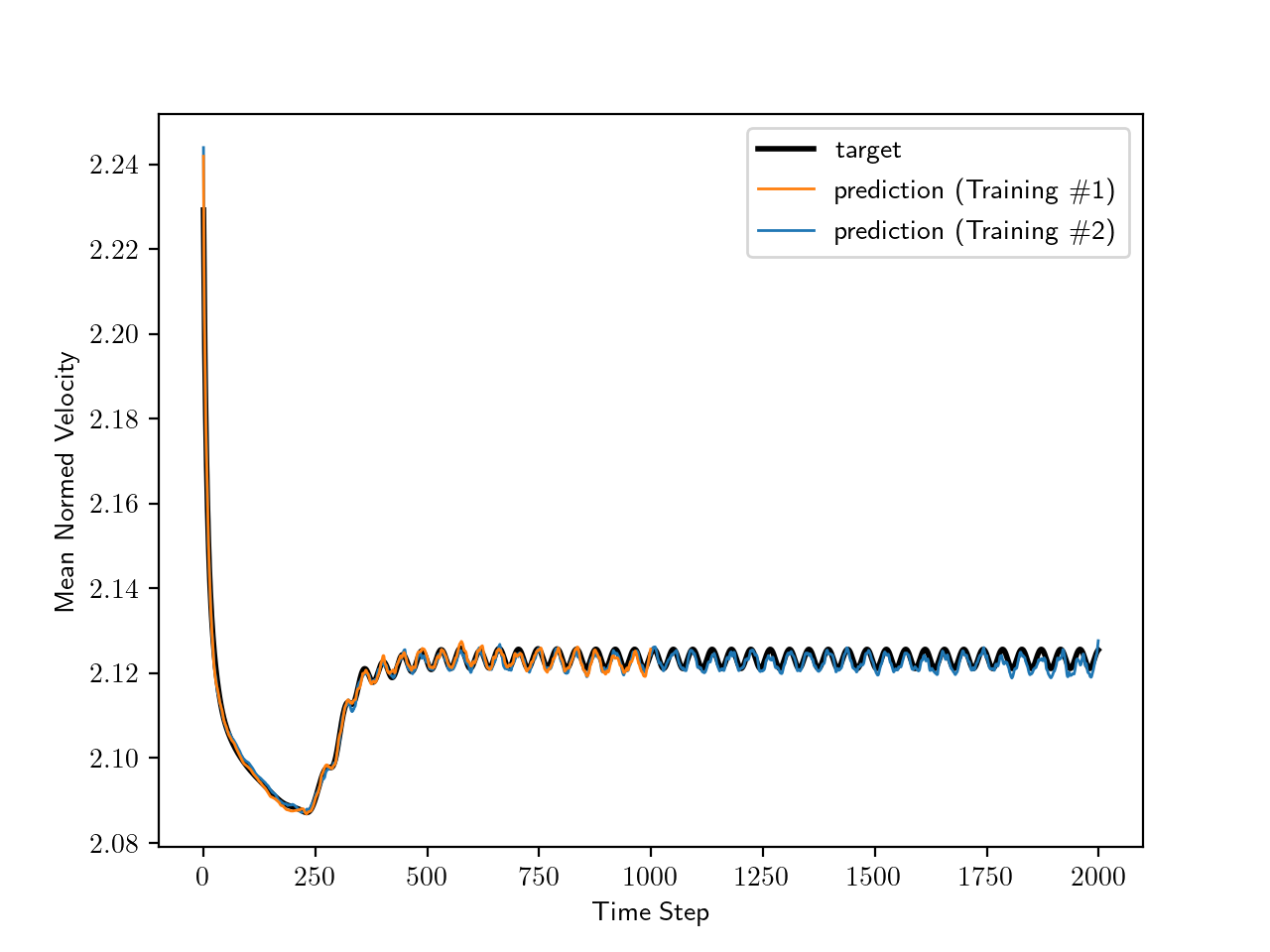}
    \caption{Spatially averaged normed velocities}
    \label{fig:vk_velocity}
  \end{subfigure}
\begin{subfigure}[c]{0.45\textwidth}
  \centering
  \includegraphics[width=\textwidth]{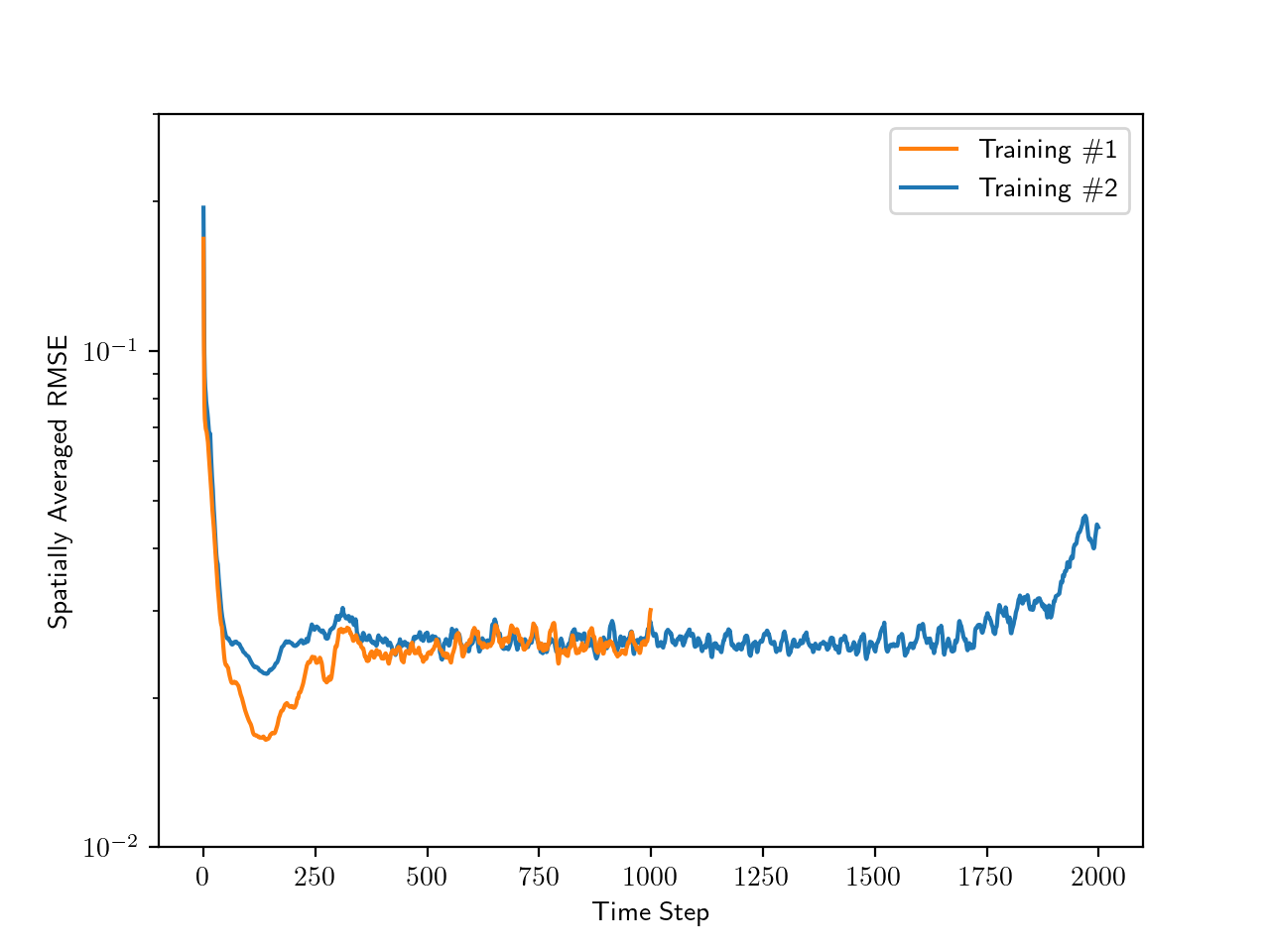}
  \caption{Full trajectory RMSE}
  \label{fig:rmse_von_karman}
\end{subfigure}
  \begin{subfigure}[b]{0.45\textwidth}
    \centering
    \includegraphics[width=\textwidth]{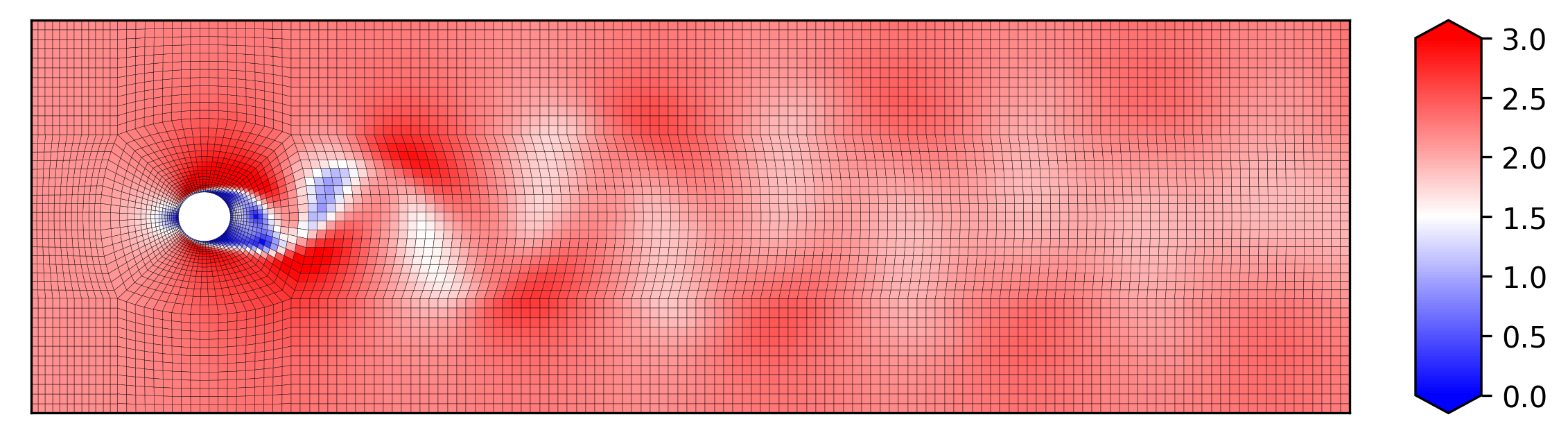}
    \caption{Normed velocity target at time step 1000}
    \label{fig:vk_target}
  \end{subfigure}
   \begin{subfigure}[b]{0.45\textwidth}
    \centering
    \includegraphics[width=\textwidth]{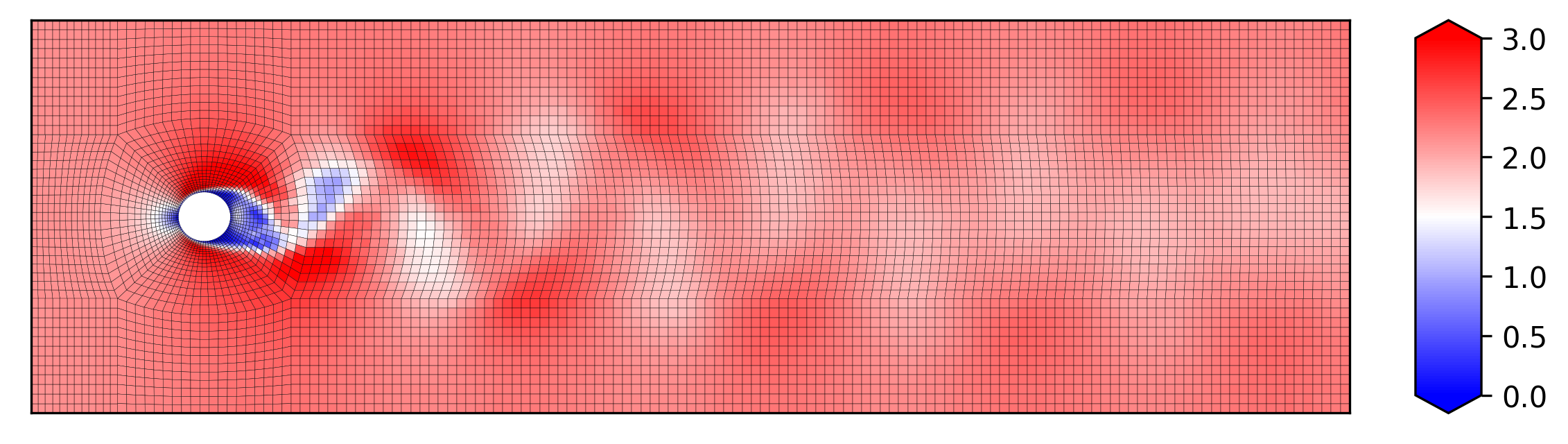}
    \caption{Normed velocity prediction at time step 1000}
    \label{fig:vk_pred}
  \end{subfigure}

\caption{Generalization test: (\ref{fig:vk_velocity}) compare the average normed
velocity for prediction and target. (\ref{fig:rmse_von_karman}) displays the
RMSE for the full trajectory. Training \#1 is performed with the first 1000 time steps only, training \#2  with 2000 time steps.
(\ref{fig:vk_target}) and (\ref{fig:vk_pred}) give a snapshot of target
and prediction at time step 1000 for training \#2.}

  \label{fig:vk_results}

\end{figure}

We tested our approach on two canonical CFD examples, the lid-driven cavity and
the von K\'arm\'an vortex street. For the sake of conciseness, here we only
present results for the most complex case. The von K\'arm\'an vortex street
corresponds to a fluid passing a cylinder with imposed input velocity.  Our
architecture is implemented with pytorch-geometric \cite{fey2019fast}, with 6
spline convolutions connecting 7  meshes of  increasing resolution. The meshes
are regularly covering the domain $\Omega$ with $h_{1 \leq k \leq 6} = \lceil
\frac{l}{2^{4-k}} \rceil$, where $l$ is the length of $\Omega$. The spline
convolutions are of degree 1, with 3 points of control for each of the 2
dimensions. The numbers of output features for the 6 spline convolutions are
respectively 512, 256, 128, 64, 32 and 3. The 3 final outputs correspond to the
scalar pressure field and the 2-dimensional vector velocity field. The size of
the network (34MB for 8.5M parameters) is considerably more compact than the
training data. We used  the Code-Saturne solver \cite{archambeau2004code} to
generate a dataset of 20 simulations (2GB per simulation). The 20 simulations
ran with an irregular mesh of 7361 cells consisting of 15176 points, with an
input velocity randomly sampled between 2 and 2.2 (Reynolds numbers
between 3000 and 3300). Each simulation runs for 2000 time steps. The model
inputs are thus $t \in [0, 2000]$ and $\lambda \in [2, 2.2]$. Train,
validation, and test splits account respectively for 18, 1, and 1 simulations.
Each simulation is normalized pointwise. We computed the mean and standard
deviation of each point across the different time steps and set of parameters.
Training took  1000 epochs using Adam optimizer. Initial learning rate was set
to 5E-2 and decreased of factor 0.1 with a reduce on plateau scheduler of
patience 10. The minimum learning rate was 1E-6.

\hidden{
\paragraph{Lid-driven cavity} The lid-driven cavity consists in an incompressible fluid contained in a square
domain. We impose Dirichlet boundary conditions. On the top left corner the
velocity is imposed to be $u_0$ and null elsewhere. In the domain, the fluid is
initially at rest. For this problem, the inputs of the model are the time step
$t$ of the simulation, the velocity $u_0$, and the viscosity of the fluid $\mu$.
The 20 simulations of the dataset count 300 time steps on a mesh of 127 nodes.
The input velocity $u_0$ is randomly sampled between 0 and 7. The viscosity is
randomly sampled between 5E-4 and 2E-3 covering common values of the water
viscosity at different temperatures. We obtain an error RMSE of 1E-4 on the test
simulation. Figure \ref{fig:ldc_pred} shows predictions accurately reproduce
target time step.
}

Although never seen
during training the test simulation belongs to the same distribution as the
training dataset. Results shown here are thus interpolation both in time and
velocity space. On the test simulation (Fig.\ref{fig:vk_results}), the error averages 1.2E-2 with a standard deviation of 2.7E-2, similar to what was obtained during training.
Generalization from training \#2 well matches the target without an increasing
error accumulation along the trajectory, except for the last time steps (above
1750). To further understand this  phenomena, unexpected for a direct time
approach, we performed a second training with only the 1000 first time steps
(training \#1). A similar increase of the error is visible when getting close to
the last time steps. This suggests the observed increase is related to the lack
of training data points surrounding the time horizon.
\section{Conclusion}
\label{sec:conclusion}

In this article we have proposed a novel deep surrogate architecture for
Computational Fluid Dynamics. It differs from the state-of-the-art by its
capability to compute the fluid state directly  and to support irregular space
discretizations. Early experiments show encouraging results with a
generalization RMSE  that is both low  and not subject to error accumulation. So
far, training and generalization were tested within  a rather tight range of
varying parameters. Future work will focus on testing and refining the presented
architecture for a wider spectrum of simulation parameters. \hidden{ \lucas{The practicality of
the network for parameter space exploration would be corroborated by maintaining
similar performances on wider range of Reynolds numbers.}}

\printbibliography


\end{document}